\documentclass[11pt]{article}

\usepackage[preprint]{acl}

\usepackage{times}
\usepackage{latexsym}

\usepackage[T1]{fontenc}

\usepackage[utf8]{inputenc}

\usepackage{microtype}

\usepackage{inconsolata}

\usepackage{graphicx}
\usepackage{booktabs}
\usepackage{amsmath,amssymb}
\usepackage{subcaption}
\usepackage{adjustbox}
\usepackage{multirow}
\usepackage{newunicodechar}
\newunicodechar{Δ}{$\Delta$}

\usepackage{algorithm}
\usepackage{algorithmic}
\usepackage{listings}
\usepackage{seqsplit}
\usepackage{xurl}
\usepackage{enumitem}
\usepackage{longtable}

\usepackage[most]{tcolorbox}
\tcbuselibrary{listings,breakable}
\usepackage{tabularx,array}

\title{When Reasoning Supervision Hurts: TTCW-Based Long-Form Literary Review Generation}

\author{
  Jinlong Liu \quad Mohammed Bahja \quad Mark Lee \\
  School of Computer Science, University of Birmingham, United Kingdom \\
  \texttt{jxl2069@student.bham.ac.uk;\{m.bahja,m.g.lee\}@bham.ac.uk}
}

\begin{document}
\maketitle
\begin{abstract}
Automatic evaluation of long-form literary writing remains challenging, particularly for creativity-related dimensions such as originality and flexibility. We address the lack of large-scale supervision for structured Torrance Test of Creative Writing (TTCW)-based evaluation by constructing a dataset of 263{,}911 long-form stories, each paired with scalar scores and meta-synthesised review comments across 14 TTCW-based dimensions. Using this dataset, we study whether reviewer-style reasoning traces improve structured literary review generation across Qwen3-4B, Qwen3-8B, and Nemotron-8B. Across all three models, no-trace supervision yields higher score imitation fidelity and near-complete report generation, whereas trace supervision reduces completion to $85.9$--$90.9\%$. Failure analysis shows that $88.6\%$ of failed reasoning generations contain semantic repetition loops inside the trace. These loops concentrate in particular story--criterion interactions and progressively harden after onset, while comparable repetition is rare in the supervision data. Our results suggest that, for long-form tasks requiring strictly structured outputs, explicit reasoning traces can introduce generation instability that outweighs their overall benefit\footnote{Code available at \url{https://github.com/Vince-Liuss/TTCW-based-Review}}.
\end{abstract}

\section{Introduction}
In recent years, LLM-as-Judge has become increasingly common and has shown promising reliability across multiple evaluation settings \citep{bonomo-etal-2025-literaryqa,chiang-lee-2023-large,liu-etal-2023-g}. At the same time, more benchmarks and resources have been proposed for long-form literary or narrative evaluation, including \textsc{ABSEval} \citep{liang-etal-2024-abseval}, \textsc{STORYWARS} \citep{du-chilton-2023-storywars}, and \textsc{CollabStory} \citep{venkatraman-etal-2025-collabstory}. In parallel, \citet{10.1145/3613904.3642731} introduce TTCW as a structured creativity-oriented evaluation framework, and \citet{li-etal-2025-automated-creativity} further propose a reference-based TTCW evaluator.

However, current work still lacks a public dataset for long-form TTCW-based review generation in a reference-free setting. Existing long-form literary evaluation resources do not provide TTCW-grounded review reports as supervision, and existing TTCW-based evaluation work has not released a large-scale dataset focused on literary review generation. This leaves a gap for training judge-style models that must produce both metric-aligned scores and review comments under a structured rubric.

To address this gap, we construct a large TTCW-based literary review dataset by converting the original TTCW binary questions into scalar rating questions from 1 to 10. We ask three reviewer models to score all 14 TTCW metrics independently for each story, evaluate reviewer quality through score distribution, discrimination, and metric-isolation analyses, remove the weakest reviewer, and then use a separate model to synthesise the remaining metric-wise comments into final review reports. The resulting dataset contains 263,911 rows of long-form stories in the 4K--8K word range, each paired with a complete TTCW-based review report.

Using this dataset, we further study whether reasoning supervision improves performance on this structured rubric-based review task. We compare models fine-tuned with and without reasoning content, and find that the non-reasoning setting performs better overall. The results suggest that, for fixed-format review generation with explicit score prediction, reasoning content does not improve performance and may instead reduce output stability.
Our main contributions are as follows:
\begin{itemize}
    \item We construct a large TTCW-based dataset for long-form literary review generation by converting the original binary TTCW questions into scalar rating-based review supervision.
    \item We design a dataset construction pipeline that performs metric-wise reviewer scoring, reviewer-quality filtering, and comment synthesis to produce complete TTCW review reports for long-form stories.
    \item We provide an empirical comparison of reasoning and non-reasoning fine-tuning on this structured review task, and show that non-reasoning supervision performs better in our setting.
\end{itemize}

\section{Related Work}
\label{sec:related}
We review two strands: (i) \emph{LLM-as-Judge} for open-ended text evaluation, and (ii) \emph{long-form literature} resources and metrics from the past two years. We then identify a supervision gap around the TTCW.

\subsection{LLM-as-Judge}
\citet{bonomo-etal-2025-literaryqa} introduce \textsc{LiteraryQA}, a cleaned subset of NarrativeQA focused on literary works, and conduct a meta-evaluation showing that n-gram metrics correlate weakly with human judgments, whereas LLM judges—including small open-weight models—recover human-like rankings under a reference-based protocol. 
\citet{chiang-lee-2023-large} evaluate “LLM-as-evaluator” by giving models the same instructions and items used in human studies; model ratings track expert judgments and remain stable across prompt formatting and sampling choices.
\citet{liu-etal-2023-g} propose G-EVAL, where GPT-4 as the judge achieves Spearman $\rho=0.514$ with human on summarization, illustrating that rubric-prompted judging can reach competitive human alignment.

Discourse-level analyses highlight where generic judges may miss narrative structure. \citet{tian-etal-2024-large-language} analyze story arcs, turning points, and affect; baseline arc identification is near random for mid-tier models, improves for frontier models, but remains below human; explicitly modeling arcs/affect boosts narrative diversity, suspense, and arousal.

TTCW operationalizes creativity as a product via 14 binary tests across Fluency, Flexibility, Originality, and Elaboration \citep{10.1145/3613904.3642731}. Reported per-test interrater agreement is moderate, while aggregate agreement is strong, supporting TTCW as a reproducible \emph{set}-based evaluation protocol. Recent surveys catalog limitations of LLM-as-Judge (e.g., sentiment, token, and context/culture biases) and outline reliability practices (e.g., pairwise comparisons, bias controls).

\subsection{Long-Form Literature Resources and Metrics}
\textbf{Scripted and collaborative narratives.}
\citet{liang-etal-2024-abseval} propose \textsc{ABSEval} with \textsc{MCScript} (1{,}500 tasks) and report closer alignment with human judgments than single-LLM setups; top systems include strong chat models, and the agentic framework improves agreement with human evaluators. 
\citet{du-chilton-2023-storywars} release \textsc{STORYWARS} (40k human-authored collaborative stories; 12 task types, 101 tasks). 
\citet{venkatraman-etal-2025-collabstory} build \textsc{CollabStory} (32k LLM-coauthored stories) and show that standard baselines struggle on authorship-related tasks; fine-tuned Transformers perform strongly on boundary authorship verification.

\textbf{Character cognition and inner thought.}
\citet{xu-etal-2025-guess} present \textsc{ROLETHINK} (6{,}058 instances from 76 books) for character-thought generation; MIRROR (memory retrieval + chain-of-thought) outperforms baselines. Gold (original monologues) is harder than silver (expert analyses), indicating sensitivity to reference fidelity and memory access.

\textbf{Long-context generation and long-text modeling.}
\citet{liu-etal-2024-longgenbench} introduce \textsc{LongGenBench} for long-context \emph{generation} (logical flow); higher-baseline models degrade less, and within-series scaling (e.g., LLaMA-3, Qwen2) reduces the performance drop. 
\citet{guan-etal-2022-lot} propose \textsc{LOT} (Chinese long text) and show that \textsc{LongLM} pretrained on 120G novels substantially outperforms similar-sized baselines on understanding and generation, with high agreement for human-labeled understanding tasks.
\citet{yang-jin-2025-matters} introduce \textsc{LongStoryEval} (600 books; avg.\ 121k tokens), derive aspect criteria from reader critiques, and report that \textsc{NovelCritique} aligns best with human ratings overall and on most aspects.

\textbf{Stress tests and judge models.}
\citet{he-etal-2023-blind} design synthetic stress tests that expose blind spots in model-based metrics, recommending metric combinations and robustness probes. 
Judge models fine-tuned for evaluation include \textsc{PandaLM}, which recovers a large fraction of GPT-3.5/4’s evaluation ability on its testbed and improves base models under its tuning regimen \citep{wang2024pandalmautomaticevaluationbenchmark}, and \textsc{Themis}, a reference-free evaluator trained with consistency verification and rating-oriented preference alignment, reporting the best average performance across six Natural Language Generation(NLG) tasks in its setup \citep{hu-etal-2024-themis}.
\citet{wu2025writingbenchcomprehensivebenchmarkgenerative} present \textsc{WritingBench} (six domains, 100 subdomains) with a fine-tuned critic; some English prompts request emulation of non-English figures (e.g., “write a story as Li Bai”), which can produce translationese-like prose rather than native English literary writing and complicate cross-domain comparability.

\textbf{Creativity-targeted evaluation.}
\citet{li-etal-2025-automated-creativity} propose a reference-based TTCW evaluator and report improved alignment (pairwise accuracy up to 0.75). 
Complementary work on creative reward shaping (RLAIF) reports strong agreement with human judgments in constrained creative settings (e.g., Chinese greetings) and underscores the role of principled judge prompts or reward models \citep{wei-etal-2025-igniting}. 
Bias analyses for complex evaluation contexts find auxiliary-information-induced vulnerabilities in LLM judges, motivating explicit robustness checks \citep{li-etal-2025-curse}.

\subsection{Gap: Supervision for TTCW-Grounded Evaluation}
Despite recent progress, there is still no public long-form dataset with TTCW-labelled supervision for automated judges. Existing judge models are typically trained on generic rubrics, while long-form literary benchmarks do not provide TTCW-grounded review supervision. As a result, current evaluation settings may capture surface quality more easily than creativity-related dimensions such as originality and flexibility. We address this gap by constructing a TTCW-based dataset for long-form literary review generation and using it to study structured rubric-based evaluation.

\begin{table*}[t]
\centering
\scriptsize
\setlength{\tabcolsep}{4pt}
\renewcommand{\arraystretch}{1.05}

\begin{tabularx}{\textwidth}{
    @{}
    >{\raggedright\arraybackslash\bfseries}p{0.15\textwidth}
    >{\raggedright\arraybackslash}X
    @{}
}
\toprule

Shared System Prompt &
You are an experienced fiction editor preparing manuscripts for publication.
You evaluate writing like you would for a serious author revising toward publication.

\textbf{General rules:}
You judge ONLY the specific craft metric you are given.
You must ground every judgment in concrete evidence from the story
(scenes, passages, beats). Do not make abstract claims with no textual anchor.
You must give both strengths (what should be preserved) and revision priorities
(what should change first). Editors do both.
You must assign a score using the rubric below. You are allowed to use any
integer 1--10. Do not collapse to the middle just to be ``safe.'' If the work
is excellent for this metric, score high. If it is weak, score low. This
instruction overrides any instinct to hedge.

\textbf{Scoring rubric (used for ANY metric):}
10 = Publication-ready control of this metric. Consistent, intentional,
supports the story's emotional/structural goals.
8 = Strong. The metric is working in most places; only light refinements needed.
6 = Mixed. The core skill is present but unreliable. Some scenes undercut the
intended effect. Needs targeted revision.
4 = Weak. The metric frequently misfires or distracts. Multiple sections need
significant rewrite.
2 = Fundamentally not working. The intended effect is mostly lost for this metric.
1 = Essentially absent or actively damaging the story.
You may still choose any integer 1--10. The descriptions are anchors, not the
only valid scores.

\textbf{Required output format:}
\texttt{Reasons: [The detailed reasoning you used to arrive at your score,
including specific examples from the story]}
\newline
\texttt{Score: [single integer 1-10]}
\newline
You must follow that exact formatting.
\\

\midrule

Metric Prompt:
Fluency1
(Narrative Pacing) &
\texttt{\{story\}}

\textbf{\#\#\# Expanded Expert Measure}

`Compression/stretching of time' in fiction writing, also known as pacing,
refers to the manipulation of time in storytelling for dramatic effect,
pacing, or other narrative purposes. Essentially, it's about controlling
the perceived speed and rhythm at which a story unfolds. Compression of time
refers to when events that take a long time (hours, days, weeks, or even years)
are summarized or condensed into a brief narrative span. For example, a writer
might compress several years of a character's life into a few paragraphs to
quickly convey important changes or developments.

On the other hand, stretching of time is when a brief moment or event is drawn
out over pages or chapters. It's often used to create suspense, emphasize
details, or delve deeper into a character's thoughts and feelings. For example,
the few seconds it takes for a dropped glass to hit the floor might be stretched
out with detailed descriptions of the action, reactions, and thoughts of
characters involved.

Storytime refers to the time within the world of the story, while real-world
time refers to the time it takes for the reader to read the story. A skilled
writer can manipulate the relationship between these two to affect the pacing
of the narrative, either speeding it up (compression) or slowing it down
(stretching). This technique plays a crucial role in shaping the reader's
experience and engagement with the story.

Given the story above, list out the scenes in the story in which time compression
or time stretching is used, and argue for each whether it is successfully
implemented. Then overall, give your reasoning about the question below and
give an answer from 1 to 10, where 10 is the best score and 1 is the worst score.

\textbf{Q)} How appropriate and balanced does the manipulation of time in terms
of compression or stretching feel?
\\

\bottomrule
\end{tabularx}

\caption{Full prompts used for dataset construction. The shared system prompt
is applied to all TTCW criteria, while Fluency1 shows the complete
Narrative Pacing prompt. The \texttt{\{story\}} placeholder is replaced by the
full input story. The expanded criterion definition and evaluation question
follow \citet{10.1145/3613904.3642731}.}
\label{tab:prompt_fluency1}
\end{table*}

\section{Dataset Preparation}
\label{sec:dataset}

We first reformulate the original TTCW metric questions from binary judgments to scalar ratings on a 1--10 scale, embedding explicit score anchors in the system instruction so that all reviewer models operate under the same rubric, shown as Table \ref{tab:prompt_fluency1}. To minimise cross-metric interference and reduce the risk of a reviewer collapsing multiple criteria into a single latent judgment, we evaluate the 14 metrics independently rather than jointly; the full metric list is provided in the Appendix.

We select three recent and capable reviewer models: \texttt{\seqsplit{Llama-3\_3-Nemotron-Super-49B-v1\_5}} \citep{bercovich2025llamanemotronefficientreasoningmodels}, \texttt{\seqsplit{Qwen3-Next-80B-A3B-Instruct}} \citep{qwen3technicalreport} (non-reasoning mode), and \texttt{\seqsplit{gpt-oss-120b}} \citep{openai2025gptoss120bgptoss20bmodel}. For source fiction, we use the \texttt{\seqsplit{WritingPrompts}} corpus \citep{fan-etal-2018-hierarchical}. Because many stories fall below the length threshold suitable for long-form evaluation, we remove samples exceeding 8K words and use \texttt{\seqsplit{Gemma-3-27b-it}} \citep{geminiteam2025geminifamilyhighlycapable} to regenerate stories from the original prompts, treating human-written stories as references, to obtain samples in the 4K--8K word range. Each reviewer evaluates stories independently by metric, while GLM-4.5-Air \citep{5team2025glm45agenticreasoningcoding} synthesises the two retained reviews, prioritising shared strengths and weaknesses while acknowledging disagreements. All models use $T=0$.

\begin{figure}[h]
    \centering
    \includegraphics[width=0.9\linewidth]{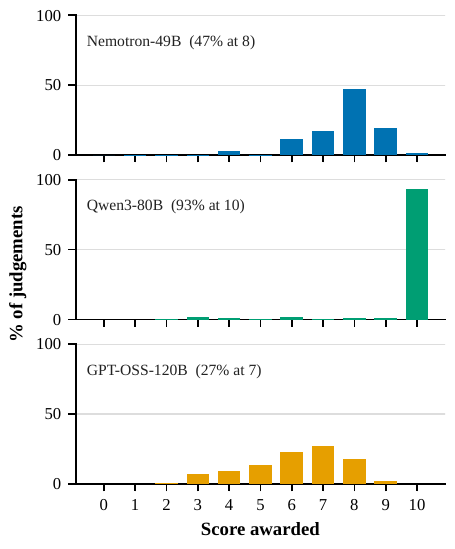}
    \caption{Score distributions of the three LLM reviewers aggregated across the 14 TTCW metrics. Nemotron-49B has a mode of 8 (47\%), Qwen3-80B a mode of 10 (93\%), and GPT-OSS-120B a mode of 7 (27\%).}
    \label{fig:score_distribution}
\end{figure}

\begin{figure}[h]
    \centering
    \includegraphics[width=0.8\linewidth]{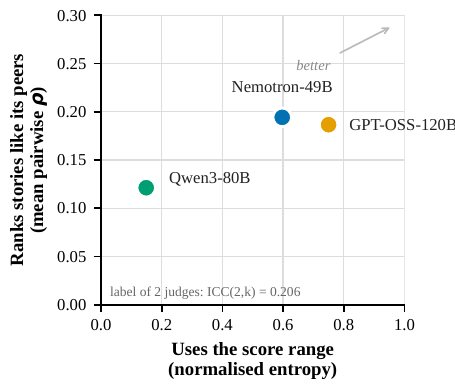}
    \caption{Reviewer discrimination and ranking agreement. The x-axis measures score-range utilisation using normalised entropy, and the y-axis measures mean pairwise Spearman correlation ($\rho$) in story ranking; higher values are preferred on both dimensions. The retained two-reviewer agreement is ICC$(2,k)=0.206$.}
    \label{fig:judge_quality}
\end{figure}

\begin{figure*}[h]
    \centering
    \includegraphics[width=0.75\linewidth]{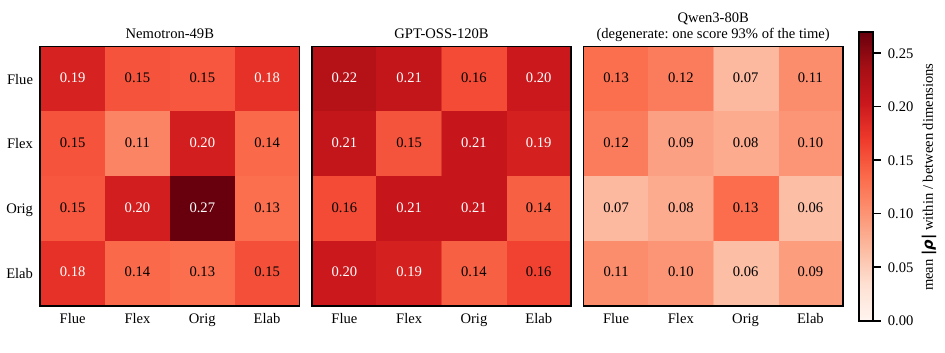}
    \caption{Group-level inter-metric correlations across reviewer models. The 14 TTCW metrics are aggregated into Fluency, Flexibility, Originality, and Elaboration; diagonal cells show mean within-dimension correlation and off-diagonal cells mean cross-dimension correlation. Lower correlations for Qwen3-80B should be interpreted in light of its limited score variation. Full heatmaps are in Appendix Fig.~\ref{fig:heatmaps_a}.}
    \label{fig:compacted_heatmap}
\end{figure*}

Before finalising the dataset, we assess reviewer suitability using three diagnostics: \textit{score distribution}, which detects score concentration and ceiling effects; \textit{discrimination}, which measures whether a reviewer uses the score scale sufficiently to distinguish among stories; and \textit{metric isolation}, which examines whether the 14 TTCW metrics are treated as distinct criteria rather than collapsed into a single latent quality judgement. The main results are summarised in Fig.~\ref{fig:score_distribution}, Fig.~\ref{fig:judge_quality}, and Fig.~\ref{fig:compacted_heatmap}. Detailed per-metric score distributions and inter-metric correlation heatmaps are provided in Appendix Fig.~\ref{fig:overall_distribution} and Fig.~\ref{fig:heatmaps_a}, respectively.

Compared with \texttt{\seqsplit{gpt-oss-120b}} and \texttt{\seqsplit{Llama-3\_3-Nemotron-Super-49B-v1\_5}}, \texttt{\seqsplit{Qwen3-Next-80B-A3B-Instruct}} shows weaker reviewer suitability. As shown in Fig.~\ref{fig:score_distribution}, its scores are overwhelmingly concentrated at 10, indicating a pronounced ceiling effect and little practical differentiation among stories. Fig.~\ref{fig:judge_quality} provides a complementary view: Qwen3-80B has both the lowest normalised score entropy and the lowest mean pairwise ranking correlation. In contrast, GPT-OSS-120B uses the score range most broadly, while Nemotron-49B achieves the highest ranking agreement with substantially greater score variation than Qwen3-80B.

Metric isolation provides a separate diagnostic. Although Fig.~\ref{fig:compacted_heatmap} shows lower inter-metric correlations for Qwen3-80B, these correlations are not sufficient evidence of better metric isolation because the model exhibits very limited variation in its score assignments. We therefore interpret its correlation structure jointly with the score-distribution and discrimination diagnostics. Taken together, the three diagnostics indicate weaker reviewer suitability for Qwen3-80B.

We therefore exclude \texttt{\seqsplit{Qwen3-Next-80B-A3B-Instruct}} from the synthesis stage and retain the remaining two reviewer models. The final dataset contains 263{,}911 rows and targets long-context literary review generation. Each input story contains 4K--8K words, and each output contains a complete TTCW-based review report, resulting in substantially longer sequences than standard short-form evaluation settings. For fine-tuning, the non-reasoning version uses a maximum context length of 16{,}384 tokens, whereas the reasoning version requires 32{,}768 tokens because it additionally includes reviewer-style reasoning traces\footnote{Dataset available at \url{https://huggingface.co/datasets/VibrantVista/TTCW-Based-Review}}.

\paragraph{Sample Validation.}
To further assess the quality of the meta-synthesised reviews, we conduct an automatic sample-level validation using \texttt{\seqsplit{NVIDIA-Nemotron-3-Super-120B-A12B}} \citep{nvidia_nemotron_3_2025}, a recent reasoning-oriented judge model. We randomly sample 50 stories and pair each story with its 14 metric-specific review comments, resulting in 700 story--metric review pairs. To distinguish the quality of the source reviews from that of the final synthesis, we evaluate both outputs from the two retained reviewer models and the corresponding meta-synthesised labels. For each pair, we ask the validation model three binary questions commonly used to assess NLG quality:

\begin{enumerate}
    \item \textbf{Faithfulness:} Does the review only make claims that are consistent with the story's actual content, without introducing details, events, or characterisations not present in the story?
    \item \textbf{Coherence:} Is the review logically organised and internally consistent, with no contradictory statements?
    \item \textbf{Relevance:} Does the review focus on specific aspects of this story rather than making observations that could apply to almost any story?
\end{enumerate}

The results are reported in Table~\ref{tab:sample_validation}. Both retained reviewers achieve high pass rates across all three criteria. Nemotron-49B reaches $91.2\%$ faithfulness, $96.0\%$ coherence, and $95.6\%$ relevance, while GPT-OSS-120B reaches $82.8\%$, $96.0\%$, and $98.0\%$, respectively. These results indicate that the inputs to the synthesis stage are predominantly grounded in the corresponding stories, internally coherent, and story-specific.

The meta-synthesised labels retain high relevance ($97.6\%$) and coherence ($88.6\%$), while faithfulness decreases to $72.9\%$. This indicates that the synthesis stage introduces additional noise, particularly in grounding, but does not collapse the reviews into generic or poorly organised comments. Importantly, the validation model evaluates each final review directly against its corresponding story rather than against the source reviewer outputs. The resulting pass rates therefore provide a direct estimate of the quality of the supervision used in the dataset. Although the meta-synthesised reviews should not be interpreted as human-verified gold annotations, the results support their use as automatically constructed reference supervision: most comments remain faithful to the story, nearly nine out of ten are coherent, and almost all remain specific to the corresponding story.

The reduction in faithfulness is consistent with the difficulty of the meta-synthesis step. Its input contains outputs from multiple reviewers, often with overlapping or partially inconsistent observations. The synthesis model must select useful evidence, reconcile the reviewer outputs, and organise the resulting information into a concise metric-level comment. The comparison with the source reviewers makes this degradation explicit rather than assuming that synthesis preserves reviewer quality unchanged.

\begin{table}[t]
\centering
\scriptsize
\setlength{\tabcolsep}{2.6pt}
\renewcommand{\arraystretch}{1.12}
\begin{tabular}{lccc}
\toprule
\textbf{Criterion} &
\shortstack{\textbf{Nemotron}\\\textbf{49B}} &
\shortstack{\textbf{GPT-OSS}\\\textbf{120B}} &
\textbf{Synth.} \\
\midrule
Faithfulness &
$\mathbf{91.2}\,[88.8,93.0]$ &
$82.8\,[79.8,85.4]$ &
$72.9\,[69.4,76.0]$ \\

Coherence &
$\mathbf{96.0}\,[94.3,97.2]$ &
$\mathbf{96.0}\,[94.3,97.2]$ &
$88.6\,[86.0,90.7]$ \\

Relevance &
$95.6\,[93.8,96.9]$ &
$\mathbf{98.0}\,[96.7,98.8]$ &
$97.6\,[96.1,98.5]$ \\
\bottomrule
\end{tabular}

\caption{Sample-level validation of the two retained reviewers and the meta-synthesised labels over $50$ randomly sampled stories and $700$ story--metric review pairs. Values are pass rates (\%) with $95\%$ Wilson confidence intervals in brackets; the highest pass rate for each criterion is shown in bold.}
\label{tab:sample_validation}
\end{table}

\section{Experiment}
\label{sec:experiment}

This experiment investigates whether supervising models with reviewer-style reasoning traces improves long-form literary review generation. We construct trace-supervised targets using the raw outputs of the two retained reviewer models as intermediate reasoning content, followed by the final meta-synthesised review, and compare this setting against a no-trace control trained directly on the final review output. Our main experiments use \texttt{Qwen3-8B} \citep{qwen3technicalreport}, with additional experiments on \texttt{Qwen3-4B} and \texttt{\seqsplit{Llama-3.1-Nemotron-Nano-8B-v1}} \citep{bercovich2025llamanemotronefficientreasoningmodels} to examine whether the observed behaviour generalises across model scales and families. For fine-tuning, we use LoRA with \texttt{lora\_r}$=64$ for the two 8B models and \texttt{lora\_r}$=48$ for \texttt{Qwen3-4B}, keeping the proportion of trainable parameters at approximately $2\%$ across model scales. All models use a learning rate of $2 \times 10^{-4}$ and \texttt{lora\_alpha}$=128$. For the main evaluation, we use greedy decoding ($T=0$), with alternative decoding settings examined separately in the following section.

For generation stability, we use the \textit{report completion rate}:
\begin{equation}
p_{\mathrm{comp}}
=
\frac{N_{\mathrm{complete}}}{N},
\end{equation}
where an output is counted as complete only if all 14 criterion-specific reports are successfully parsed.

To measure score imitation fidelity to the synthetic teacher, we use quadratic weighted kappa (QWK). QWK is computed separately for each criterion over completed outputs and macro-averaged across the 14 criteria. Since synthetic scores lie on a $1$--$10$ scale in half-point increments, we map them to $K=21$ categories using
\begin{equation}
c(s)=\operatorname{round}(2s).
\end{equation}
For criterion $k$, with observed count matrix $O^{(k)}$, expected count matrix $E^{(k)}$, and quadratic weights
\begin{equation}
w_{ij}
=
\frac{(i-j)^2}{(K-1)^2},
\end{equation}
we compute
\begin{equation}
\kappa_k
=
1-
\frac{\sum_{ij} w_{ij}O^{(k)}_{ij}}
     {\sum_{ij} w_{ij}E^{(k)}_{ij}},
\qquad
\mathrm{QWK}
=
\frac{1}{14}\sum_{k=1}^{14}\kappa_k.
\end{equation}
We report QWK together with the number of completed outputs used for its computation.

Review-text imitation fidelity is measured using within-criterion retrieval recall at rank 1 (R@1). Generated and reference reviews are embedded with \texttt{bge-large-en-v1.5} \citep{bge_embedding} and L2-normalised. For criterion $k$ with retrieval pool size $N_k$,
\begin{equation}
\mathrm{R@1}_k
=
\frac{1}{N_k}
\sum_{i=1}^{N_k}
\mathbb{1}
\!\left[
\arg\max_j
\mathbf{p}_i^{(k)} \cdot \mathbf{r}_j^{(k)}
= i
\right].
\end{equation}
The final score is macro-averaged across the 14 criteria:
\begin{equation}
\mathrm{R@1}
=
\frac{1}{14}
\sum_{k=1}^{14}\mathrm{R@1}_k.
\end{equation}
Retrieval is performed separately within each TTCW criterion, and we report R@1 together with its retrieval-pool size.

\begin{figure*}[ht]
    \centering
    \includegraphics[width=0.9\linewidth]{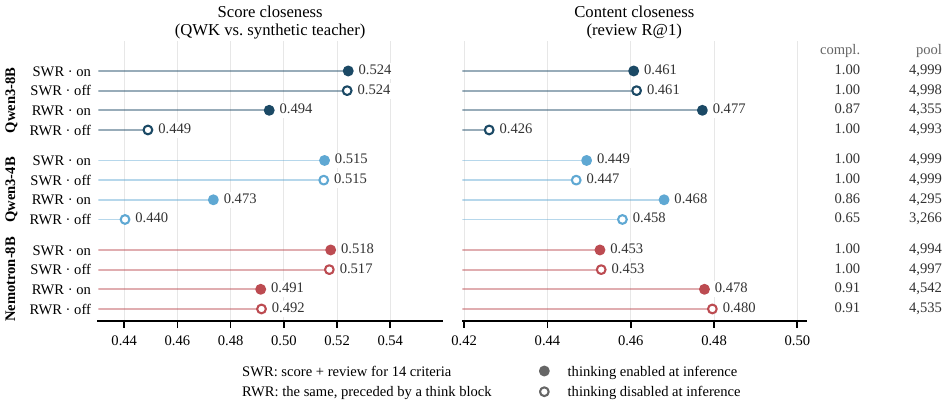}
    \caption{Performance across supervision and inference modes for \texttt{Qwen3-8B}, \texttt{Qwen3-4B}, and \texttt{Nemotron-8B}. SWR supervises the 14 criterion-level scores and reviews directly, whereas RWR prepends reviewer-style reasoning traces to the same targets. Filled and hollow markers denote thinking enabled and disabled at inference, respectively. The left panel reports score imitation fidelity (QWK), the middle panel review-text imitation fidelity (R@1), and the right columns report completion rate and the corresponding R@1 retrieval-pool size. Across all three models, SWR preserves near-complete generation and higher QWK than RWR with thinking enabled. The higher R@1 values under RWR-on are computed over smaller pools because incomplete outputs are excluded and should therefore not be interpreted as directly comparable content improvements.}
    \label{fig:metrics_overview}
\end{figure*}

The overall results are summarised in Fig.~\ref{fig:metrics_overview}, while detailed per-criterion QWK and R@1 results are reported in Appendix Table~\ref{tab:per_criterion_qwk_r1}. The clearest difference between settings is report completion. Models trained without reviewer-style traces remain highly stable across model scales and inference modes, with completion rates close to 1.0. By contrast, trace-supervised models show a substantial drop when thinking is enabled, reaching $87.1\%$ for Qwen3-8B, $85.9\%$ for Qwen3-4B, and $90.9\%$ for Nemotron-8B. Since the same 14-criterion report structure is completed reliably without trace generation, this pattern localises the instability to the intermediate reasoning process rather than the final output format. We therefore analyse the incomplete trace-supervised generations to determine what prevents them from reaching the final report.

\section{Failure Analysis}
\label{sec:failure_analysis}

\paragraph{Failure Mode.}
The dominant failure mode is a semantic repetition loop inside the \texttt{<think>} block: 3,551 of 4,007 failures ($88.6\%$) across the six reasoning runs contain such loops. These loops arise early and repeatedly regenerate short, semantically coherent review units, preventing the final report from being reached. Only $2.5\%$ correspond to conventional token-level degeneration; most instead repeat qualifications, quotations, list items, or evidence statements. Representative examples are shown in Table~\ref{tab:loop_examples}.

\begin{table}[t]
\centering
\scriptsize
\setlength{\tabcolsep}{0pt}
\renewcommand{\arraystretch}{1.05}

\begin{tabularx}{\columnwidth}{@{}Xr@{}}
\toprule
\textbf{Repeated-unit category} & \textbf{Source / repeats} \\
\midrule

\textbf{Contrastive connective}
& Nemotron-8B, 1821 \quad $\times703$ \\
\multicolumn{2}{@{}p{\columnwidth}@{}}{
``The story started with the event already in progress, \textbf{but} the narrative quickly moves through the days leading up to the event.''
} \\[3pt]

\textbf{Quotation}
& Qwen3-8B, 1978 \quad $\times404$ \\
\multicolumn{2}{@{}p{\columnwidth}@{}}{
``\textbf{Revision Priority:} The compression of time is used to establish the protagonist's emotional state. \textbf{Evidence:} \emph{The chipped ceramic of the mug warmed my hands, a small comfort\ldots}''
} \\[3pt]

\textbf{List item}
& Qwen3-8B, 3973 \quad $\times258$ \\
\multicolumn{2}{@{}p{\columnwidth}@{}}{
``\textbf{Strengths in Time Compression:} The story uses time compression to move quickly through periods of time that are not crucial to the plot. For example, the time between\ldots''
} \\[3pt]

\textbf{6d hedge}
& Qwen3-8B, 141 \quad $\times688$ \\
\multicolumn{2}{@{}p{\columnwidth}@{}}{
``This is a good use of time compression, \textbf{but it is not explicitly stated} how much time has passed. The story moves quickly from the initial confrontation to\ldots''
} \\[3pt]

\textbf{Table row}
& Qwen3-8B, 1170 \quad $\times666$ \\
\multicolumn{2}{@{}p{\columnwidth}@{}}{
``\textbf{Successful} -- reinforces. \emph{The fluff\ldots{} a dark mirror held up to the human condition.} (again) Same -- reinforces.''
} \\

\bottomrule
\end{tabularx}

\caption{Representative semantic units from reasoning-loop failures. For each content category, we show the model and row from which the example is drawn, together with the number of times the unit is repeated. Examples are selected near the median repeat count of their category and truncated for readability. Categories are not mutually exclusive.}
\label{tab:loop_examples}
\end{table}

\begin{figure*}[ht]
    \centering

    \begin{subfigure}[t]{0.32\textwidth}
        \centering
        \includegraphics[width=\linewidth]{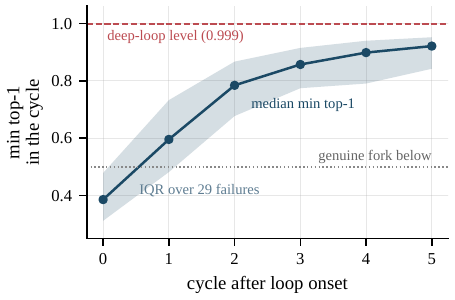}
        \caption{Loop confidence after onset.}
        \label{fig:loop_hardening_conf}
    \end{subfigure}
    \hfill
    \begin{subfigure}[t]{0.32\textwidth}
        \centering
        \includegraphics[width=\linewidth]{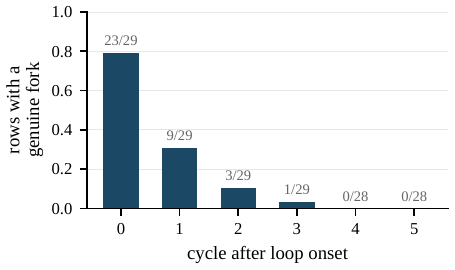}
        \caption{Remaining decoding forks.}
        \label{fig:loop_hardening_forks}
    \end{subfigure}
    \hfill
    \begin{subfigure}[t]{0.32\textwidth}
        \centering
        \includegraphics[width=\linewidth]{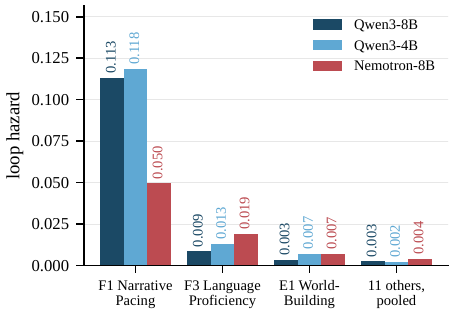}
        \caption{Loop hazard by criterion.}
        \label{fig:criterion_hazard}
    \end{subfigure}

    \caption{Failure dynamics and localisation. After loop onset, token-level confidence progressively increases while genuine decoding forks disappear; loop occurrence is also strongly concentrated in a small subset of TTCW criteria, particularly Narrative Pacing (F1).}
    \label{fig:failure_mechanism}
\end{figure*}

\paragraph{Decoding Dynamics.}
We test whether decoding affects loop entry on 200 paired Qwen3-8B stories. Increasing temperature reduces the loop rate from $14.5\%$ under greedy decoding to $4.5\%$ at $T=1.0$ ($p=0.000325$), although looping persists. However, QWK decreases from $0.4965$ to $0.4815$, and R@1 from $0.7009$ to $0.6834$, suggesting that reduced looping does not necessarily improve evaluation agreement. Suppressing a frequently observed hedge pattern has little effect ($p=0.508$). Full results are reported in Table~\ref{tab:decoding_intervention}. These findings are consistent with observations that sampling can reduce reasoning-loop failures \citep{pipis2026wait,duan2026circular}.

Once a loop begins, however, the trajectory progressively hardens. As shown in Fig.~\ref{fig:failure_mechanism}, the median minimum top-1 probability within successive loop cycles rises from $0.39$ to $0.92$, while failures retaining a genuine decoding fork fall from 23/29 to 0/28. Thus, loops are not fully deterministic at onset, but alternative continuations rapidly disappear over successive repetitions.

\begin{table}[t]
\centering
\scriptsize
\setlength{\tabcolsep}{2.5pt}
\renewcommand{\arraystretch}{1.08}
\begin{tabular}{@{}lccccc@{}}
\toprule
\textbf{Arm} &
\textbf{Loops} &
\textbf{Rec./Brk.} &
\textbf{$p$} &
\textbf{QWK} &
\textbf{R@1} \\
\midrule
Greedy ($T=0$)
& 29 (14.5\%)
& --
& --
& 0.4965
& 0.7009 \\

$T=0.6$
& 17 (8.5\%)
& 22/10
& 0.0501
& 0.4984
& 0.7013 \\

$T=1.0$
& 9 (4.5\%)
& 25/5
& 0.000325
& 0.4815
& 0.6834 \\

Hedge ban
& 26 (13.0\%)
& 6/3
& 0.508
& 0.4854
& 0.7072 \\
\bottomrule
\end{tabular}
\caption{Paired decoding interventions on 200 Qwen3-8B stories. Loops are reported as counts (rates). \textit{Rec./Brk.} denotes greedy failures recovered and complete cases broken by each intervention. $p$ reports the two-sided exact McNemar test against greedy decoding. QWK uses 155 stories completed under all conditions, while R@1
uses an equalised retrieval pool of 171.}
\label{tab:decoding_intervention}
\end{table}

\paragraph{Failure Localisation.}
Loop onset is highly concentrated across TTCW criteria. Approximately $96\%$ of loops occur in three enumeration-style criteria, with Narrative Pacing (F1) showing the highest hazard across all three models (Fig.~\ref{fig:failure_mechanism}). When two models loop on the same story, they usually fail at the same criterion: $97\%$ for Qwen3-8B/Qwen3-4B, $90\%$ for Qwen3-8B/Nemotron-8B, and $86\%$ for Qwen3-4B/Nemotron-8B, compared with marginal expectations of $77\%$, $58\%$, and $56\%$, respectively. Detailed criterion-level hazards and cross-family overlap are reported in Appendix Tables~\ref{tab:criterion_hazard} and~\ref{tab:cross_family_localisation}. This cross-family consistency suggests that loop entry is strongly associated with the interaction between the input story and criterion-specific reasoning task rather than with model family alone.

\paragraph{Supervision Audit.}
We next test whether the repetition pathology is directly inherited from the supervision data. Across all 237{,}271 training traces, only 4 ($0.0017\%$) reach the median repetition level observed in failed generations ($815\times$), compared with a $12.9\%$ generation failure rate. Moreover, completed generations remain close to the training distribution, whereas failed generations depart sharply in trace length, repetition, hedge frequency, and connective density (Appendix Table~\ref{tab:supervision_audit}). Comparable repetition is therefore too rare in the supervision data to explain the observed failure through direct imitation.

\paragraph{Value of the Trace.}
The trace nevertheless contains useful information. On the 4,350 Qwen3-8B stories completed both with and without trace generation, enabling the trace improves QWK by $+0.042$ ($95\%$ CI $[+0.034,+0.050]$). Suppressing the trace removes the deep-loop mode on Qwen3-8B, but lowers QWK below the no-trace control. The full paired decomposition is reported in Appendix Table~\ref{tab:trace_value}. Thus, reasoning traces can improve imitation fidelity conditional on successful generation, while requiring their generation introduces a stability cost that leaves the no-trace setting as the stronger overall operating point.

\paragraph{Failure Pathway.}
Together, the evidence supports a consistent failure pathway: trace generation introduces the vulnerable stage; particular story--criterion combinations are associated with loop entry across model families; and successive repetitions progressively harden the decoding trajectory until alternative continuations largely disappear. The failure is therefore not well explained by repetitive teacher traces, a single surface pattern, or one model family, but is consistent with a criterion-conditioned instability of trace generation that becomes self-reinforcing after onset.

\section{Conclusion}
\label{sec:conclusion}

We construct a 263{,}911-story TTCW-based dataset for structured long-form literary review generation and use it to examine the effect of reviewer-style reasoning supervision. Across three models, no-trace supervision consistently yields more stable generation, while trace-supervised models suffer substantial completion drops driven primarily by semantic repetition loops inside the reasoning trace. Although the traces can improve score imitation fidelity when generation succeeds, this conditional benefit is outweighed by the resulting instability. Our findings therefore suggest that, for tasks requiring long and strictly structured outputs, explicit reasoning traces may reduce generation stability rather than provide a reliable overall benefit.

\section*{Limitations}
This work has several limitations. First, dataset construction does not involve human annotators, so the supervision signal is model-generated and may contain bias, scoring noise, or synthesis errors despite the sample-level validation. Second, our experiments cover only three models from two families (\texttt{Qwen3} and \texttt{Nemotron}), which limits the generalisability of the observed failure pattern to other model families and reasoning behaviours. Third, we study only 4B and 8B models, so it remains unclear whether larger models with stronger long-context and instruction-following capabilities would exhibit the same instability. Finally, we use LoRA-based parameter-efficient fine-tuning rather than full fine-tuning, and our conclusions are specific to reviewer-style trace supervision for structured TTCW review generation.


\bibliography{anthology, custom}

\onecolumn
\appendix

\section{Additional Plots}\label{sec:plots}
\begin{figure*}[ht]
    \centering
    \begin{subfigure}[t]{0.23\textwidth}
        \centering
        \includegraphics[width=\linewidth]{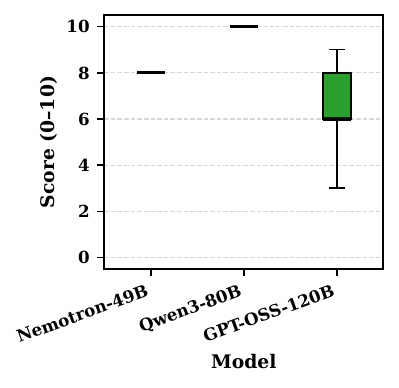}
        \caption{F1: Narrative Pacing}
        \label{fig:F1}
    \end{subfigure}\hfill
    \begin{subfigure}[t]{0.23\textwidth}
        \centering
        \includegraphics[width=\linewidth]{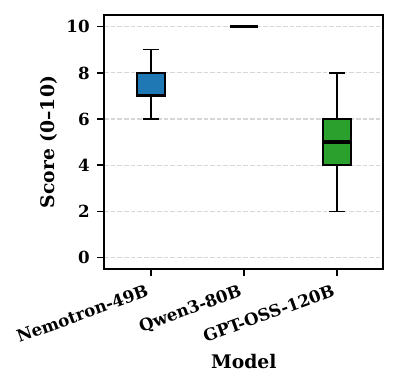}
        \caption{F2: Scene vs Exposition}
        \label{fig:F2}
    \end{subfigure}\hfill
    \begin{subfigure}[t]{0.23\textwidth}
        \centering
        \includegraphics[width=\linewidth]{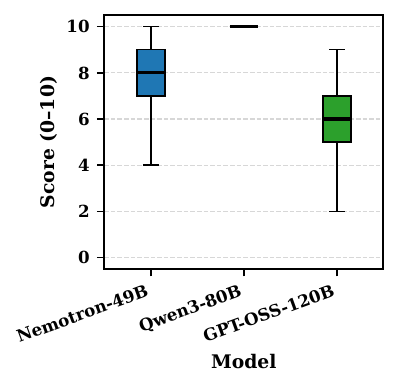}
        \caption{F3: Language Proficiency \& Literary Devices}
        \label{fig:F3}
    \end{subfigure}\hfill
    \begin{subfigure}[t]{0.23\textwidth}
        \centering
        \includegraphics[width=\linewidth]{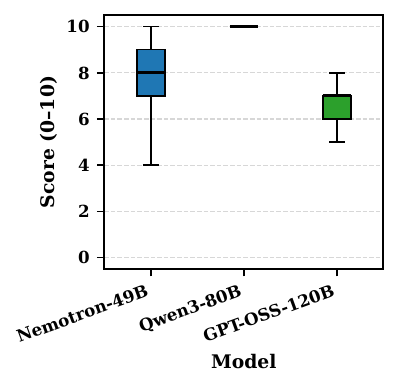}
        \caption{F4: Narrative Ending}
        \label{fig:F4}
    \end{subfigure}\hfill
    \begin{subfigure}[t]{0.23\textwidth}
        \centering
        \includegraphics[width=\linewidth]{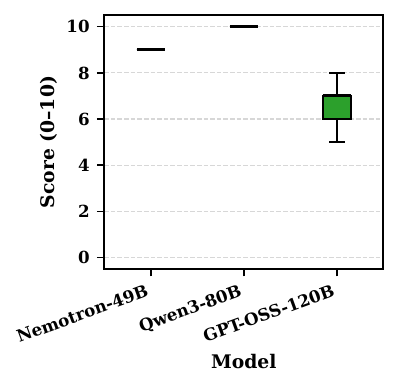}
        \caption{F5: Understandability \& Coherence}
        \label{fig:F5}
    \end{subfigure}\hfill
    \begin{subfigure}[t]{0.23\textwidth}
        \centering
        \includegraphics[width=\linewidth]{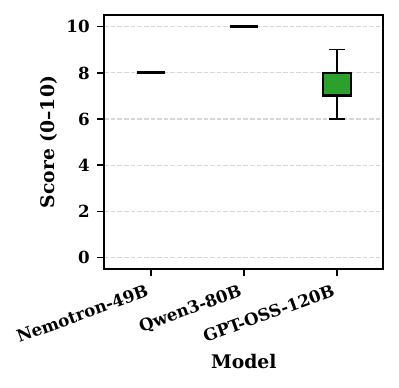}
        \caption{Fl1: Perspective \& Voice Flexibility}
        \label{fig:Fl1}
    \end{subfigure}\hfill
    \begin{subfigure}[t]{0.23\textwidth}
        \centering
        \includegraphics[width=\linewidth]{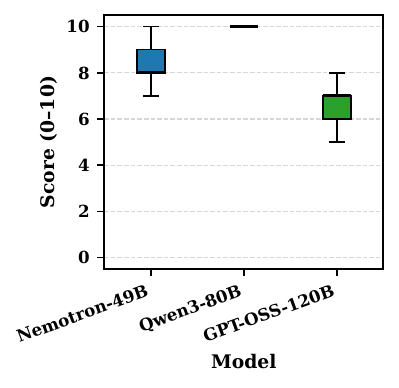}
        \caption{Fl2: Emotional Flexibility}
        \label{fig:Fl2}
    \end{subfigure}\hfill
    \begin{subfigure}[t]{0.23\textwidth}
        \centering
        \includegraphics[width=\linewidth]{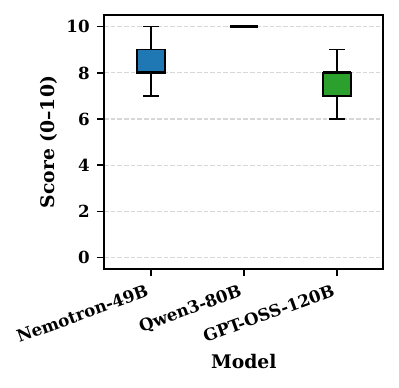}
        \caption{Fl3: Structural Flexibility}
        \label{fig:Fl3}
    \end{subfigure}\hfill
    \begin{subfigure}[t]{0.23\textwidth}
        \centering
        \includegraphics[width=\linewidth]{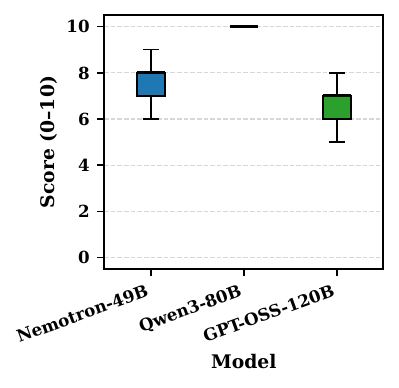}
        \caption{O1: Originality in Theme and Content}
        \label{fig:O1}
    \end{subfigure}\hfill
    \begin{subfigure}[t]{0.23\textwidth}
        \centering
        \includegraphics[width=\linewidth]{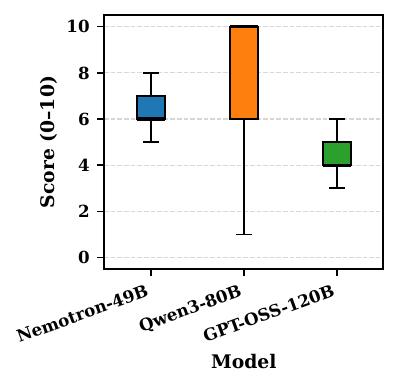}
        \caption{O2: Originality in Thought}
        \label{fig:O2}
    \end{subfigure}\hfill
    \begin{subfigure}[t]{0.23\textwidth}
        \centering
        \includegraphics[width=\linewidth]{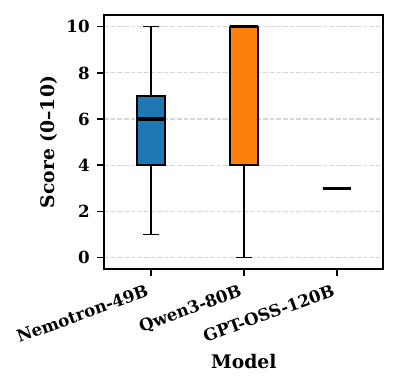}
        \caption{O3: Originality in Form}
        \label{fig:O3}
    \end{subfigure}\hfill
    \begin{subfigure}[t]{0.23\textwidth}
        \centering
        \includegraphics[width=\linewidth]{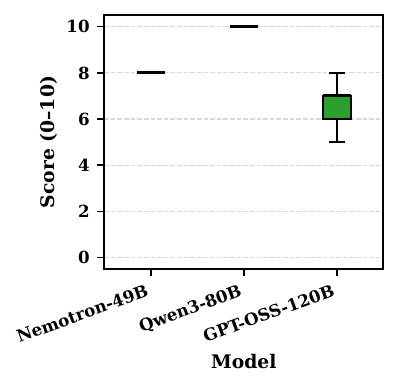}
        \caption{E1: World-Building and Setting}
        \label{fig:E1}
    \end{subfigure}\hfill
    \begin{subfigure}[t]{0.23\textwidth}
        \centering
        \includegraphics[width=\linewidth]{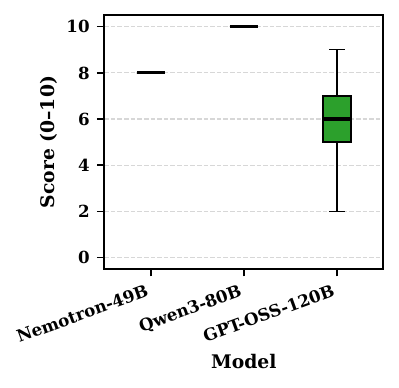}
        \caption{E2: Character Development}
        \label{fig:E2}
    \end{subfigure}\hfill
    \begin{subfigure}[t]{0.23\textwidth}
        \centering
        \includegraphics[width=\linewidth]{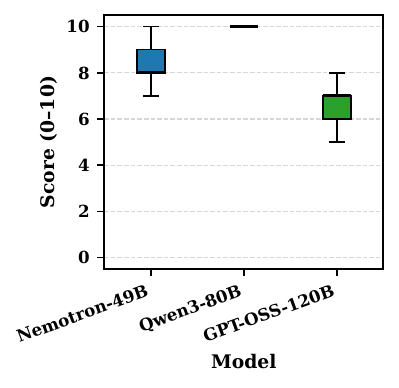}
        \caption{E3: Rhetorical Complexity}
        \label{fig:E3}
    \end{subfigure}\hfill
    \caption{Score distributions across all 14 TTCW metrics. Metric tags denote Fluency (F1--F5), Flexibility (Fl1--Fl3), Originality (O1--O3), and Elaboration (E1--E3).}
    \label{fig:overall_distribution}
\end{figure*}

\begin{figure*}[ht]
    \centering
    \begin{subfigure}[t]{0.78\textwidth}
        \centering
        \includegraphics[width=\linewidth]{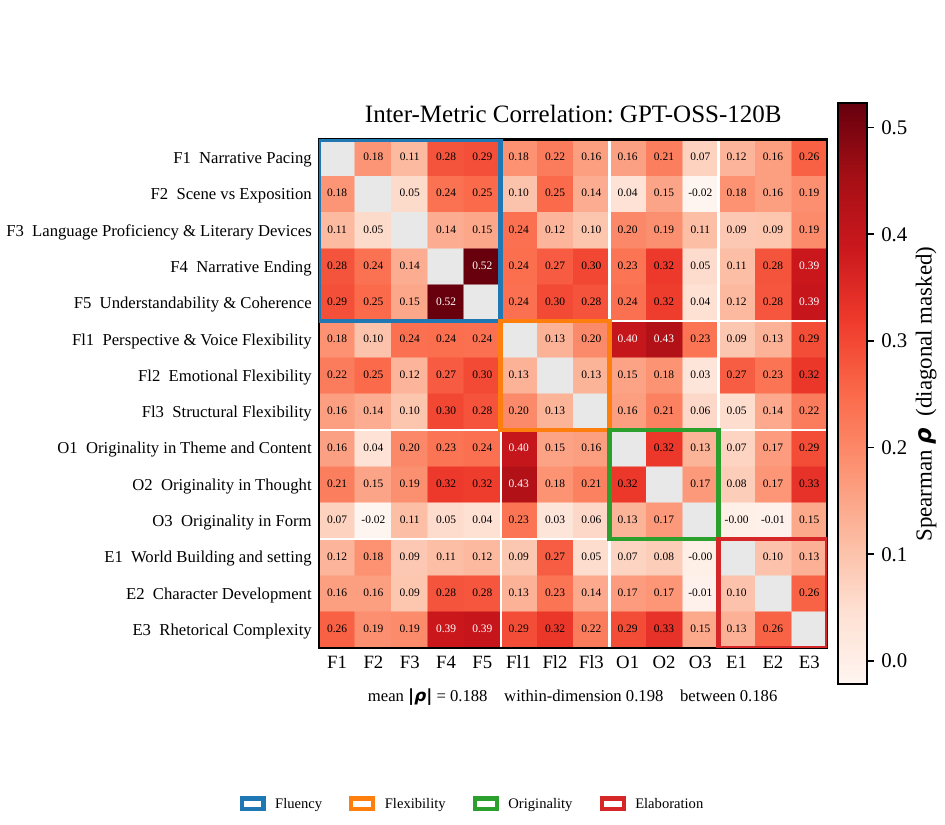}
        \caption{}
        \label{fig:heat_gpt}
    \end{subfigure}\hfill
    \begin{subfigure}[t]{0.78\textwidth}
        \centering
        \includegraphics[width=\linewidth]{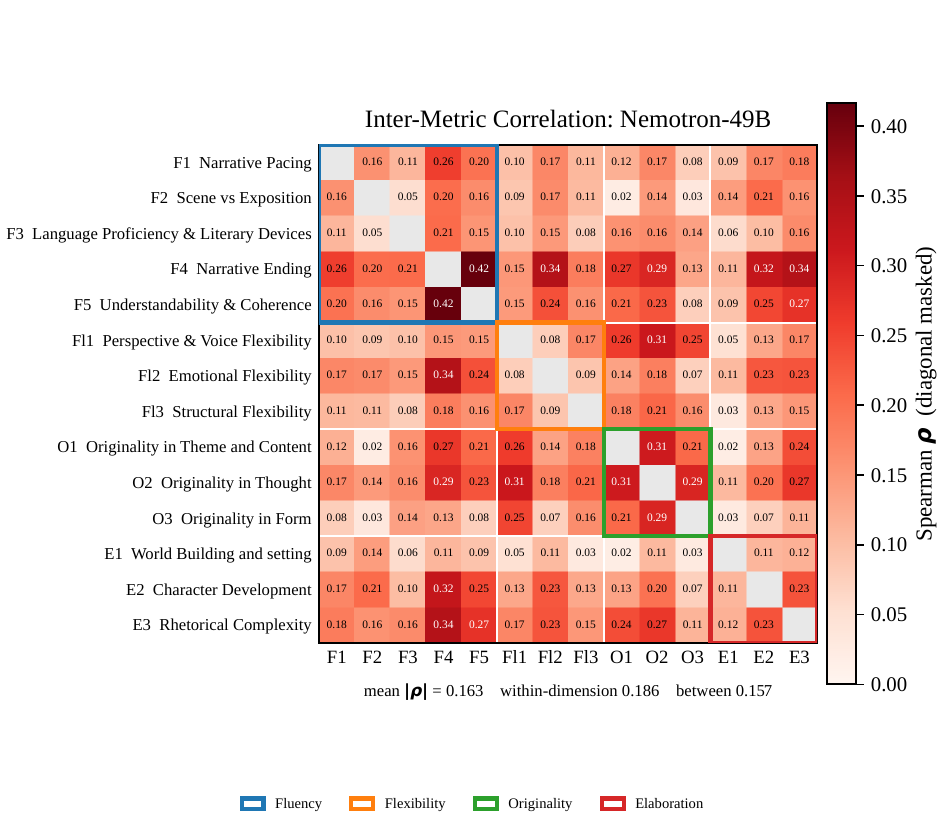}
        \caption{}
        \label{fig:heat_nemotron}
    \end{subfigure}\hfill
    \caption{Inter-metric correlation heatmaps for the three reviewer models across the 14 independently scored fiction-review dimensions. These plots diagnose whether reviewer outputs preserve metric distinctions or exhibit cross-metric coupling; stronger widespread correlations suggest greater risk of rubric collapse into broader latent quality signals.}
    \label{fig:heatmaps_a}
\end{figure*}
    
\begin{figure*}[t]
    \ContinuedFloat
    \centering
    \begin{subfigure}[t]{0.78\textwidth}
        \centering
        \includegraphics[width=\linewidth]{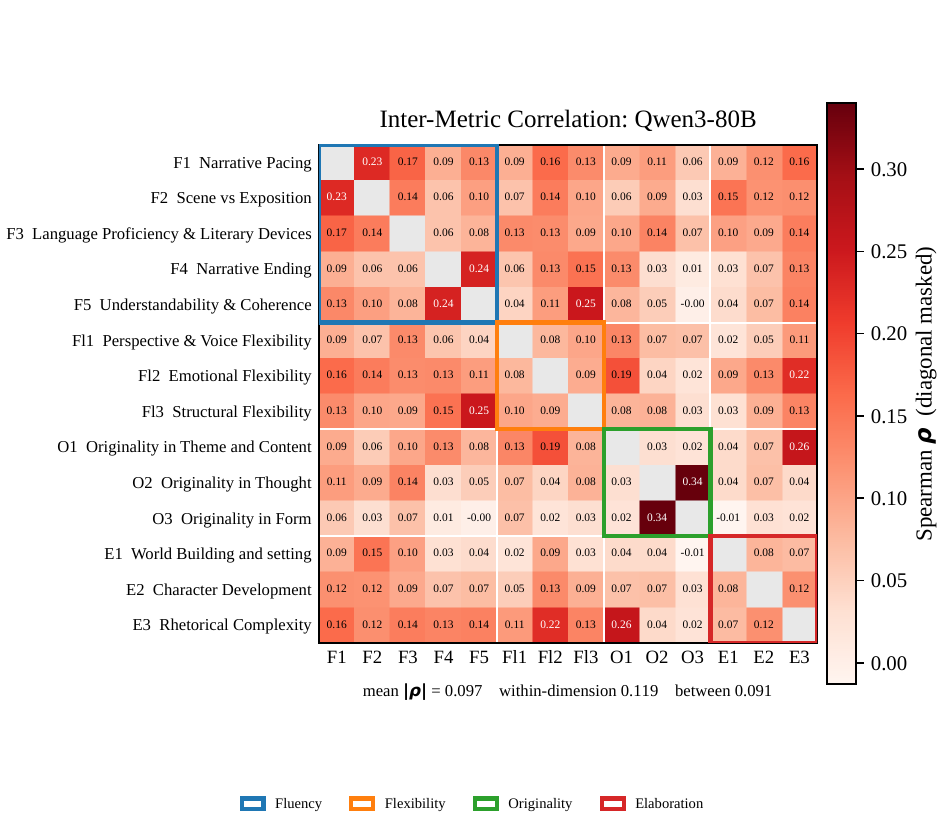}
        \caption{}
        \label{fig:heat_qwen3}
    \end{subfigure}

    \caption{Inter-metric correlation heatmaps for the three reviewer models across the 14 independently scored fiction-review dimensions. These plots diagnose whether reviewer outputs preserve metric distinctions or exhibit cross-metric coupling; stronger widespread correlations suggest greater risk of rubric collapse into broader latent quality signals.(continued)}
    \label{fig:heatmaps_b}
\end{figure*}

 \begin{table*}[ht]
  \centering
  \scriptsize
  \setlength{\tabcolsep}{1.2pt}
  \renewcommand{\arraystretch}{0.95}
  \begin{tabular*}{\textwidth}{@{}p{0.25\textwidth}@{\extracolsep{\fill}}cccccccccccc@{}}
  \toprule
   & \multicolumn{4}{c}{\textbf{Qwen3-8B}} & \multicolumn{4}{c}{\textbf{Nemotron-8B}} & \multicolumn{4}{c}{\textbf{Qwen3-4B}} \\
  \cmidrule(lr){2-5} \cmidrule(lr){6-9} \cmidrule(lr){10-13}
  \textbf{Criterion} & \multicolumn{2}{c}{No-trace} & \multicolumn{2}{c}{Trace} & \multicolumn{2}{c}{No-trace} & \multicolumn{2}{c}{Trace} & \multicolumn{2}{c}{No-trace} &
  \multicolumn{2}{c}{Trace} \\
  \cmidrule(lr){2-3} \cmidrule(lr){4-5} \cmidrule(lr){6-7} \cmidrule(lr){8-9} \cmidrule(lr){10-11} \cmidrule(lr){12-13}
   & \textbf{QWK} & \textbf{R@1} & \textbf{QWK} & \textbf{R@1} & \textbf{QWK} & \textbf{R@1} & \textbf{QWK} & \textbf{R@1} & \textbf{QWK} & \textbf{R@1} & \textbf{QWK} & \textbf{R@1}
  \\
  \midrule
  F1 Narrative Pacing
    & 0.3572 & 0.2517 & 0.2481 & 0.2707 & \textbf{0.3577} & 0.2598 & 0.2799 & 0.2620 & 0.3557 & 0.2428 & 0.2236 & 0.2584 \\
  F2 Scene vs Exposition Balance
    & 0.2881 & 0.2365 & 0.3483 & 0.2319 & 0.2819 & 0.2393 & \textbf{0.3552} & 0.2384 & 0.2797 & 0.2242 & 0.3270 & 0.2275 \\
  F3 Language Proficiency \& Literary Devices
    & 0.4074 & 0.3892 & 0.3398 & 0.3580 & 0.3944 & 0.3698 & 0.3408 & 0.3587 & \textbf{0.4089} & 0.3675 & 0.3232 & 0.3374 \\
  F4 Narrative Ending Quality
    & \textbf{0.7088} & 0.6905 & 0.6703 & 0.7104 & 0.6897 & 0.6926 & 0.6671 & 0.7182 & 0.7002 & 0.6719 & 0.6441 & 0.7141 \\
  F5 Understandability \& Coherence
    & 0.5355 & 0.6397 & 0.5020 & 0.6259 & 0.4958 & 0.6130 & \textbf{0.5469} & 0.6257 & 0.5191 & 0.6167 & 0.4980 & 0.6126 \\
  Fl1 Perspective \& Voice Flexibility
    & \textbf{0.6288} & 0.7329 & 0.5715 & 0.7662 & 0.6246 & 0.7182 & 0.5328 & 0.7609 & 0.6238 & 0.7141 & 0.5509 & 0.7546 \\
  Fl2 Emotional Flexibility
    & \textbf{0.6666} & 0.2279 & 0.5768 & 0.2595 & 0.6389 & 0.2335 & 0.5670 & 0.2508 & 0.6610 & 0.2196 & 0.5576 & 0.2475 \\
  Fl3 Structural Flexibility
    & 0.1528 & 0.3647 & 0.2289 & 0.3823 & 0.1435 & 0.3520 & \textbf{0.2459} & 0.4058 & 0.1492 & 0.3447 & 0.2205 & 0.3765 \\
  O1 Originality in Theme and Takeaway
    & 0.5865 & 0.6118 & 0.5264 & 0.6473 & \textbf{0.6046} & 0.5844 & 0.4835 & 0.6748 & 0.5797 & 0.5937 & 0.5020 & 0.6386 \\
  O2 Originality in Thought (Clich\'e Avoidance)
    & 0.5727 & 0.2571 & 0.5379 & 0.2923 & \textbf{0.5737} & 0.2401 & 0.5372 & 0.2922 & 0.5667 & 0.2446 & 0.4971 & 0.2841 \\
  O3 Originality in Form/Structure
    & 0.6556 & 0.2439 & 0.6607 & 0.2827 & 0.6512 & 0.2411 & \textbf{0.6702} & 0.2913 & 0.6294 & 0.2298 & 0.6319 & 0.2813 \\
  E1 World-Building and Sensory Believability
    & \textbf{0.7311} & 0.4076 & 0.6887 & 0.4039 & 0.7226 & 0.3900 & 0.6604 & 0.3771 & 0.7168 & 0.3977 & 0.6646 & 0.3807 \\
  E2 Character Development Depth
    & 0.4818 & 0.6775 & 0.4874 & 0.6958 & \textbf{0.5096} & 0.6730 & 0.4862 & 0.6856 & 0.4635 & 0.6681 & 0.4891 & 0.6885 \\
  E3 Rhetorical Complexity (Surface vs Subtext)
    & \textbf{0.5602} & 0.7285 & 0.5366 & 0.7532 & 0.5505 & 0.7334 & 0.5046 & 0.7466 & 0.5563 & 0.7217 & 0.4995 & 0.7506 \\
  \midrule
  Macro average (14 criteria)
    & \textbf{0.5238} & 0.4614 & 0.4945 & 0.4772 & 0.5171 & 0.4529 & 0.4913 & 0.4777 & 0.5150 & 0.4469 & 0.4735 & 0.4680 \\
  Report completion
    & \multicolumn{2}{c}{99.98\%} & \multicolumn{2}{c}{87.12\%} & \multicolumn{2}{c}{99.96\%} & \multicolumn{2}{c}{90.86\%} & \multicolumn{2}{c}{100.00\%} &
  \multicolumn{2}{c}{85.92\%} \\
  \bottomrule
  \end{tabular*}
  \caption{Per-criterion imitation fidelity to the synthetic teacher for all three models, each arm decoded in the mode it was trained for: No-trace = \texttt{score\_with\_reviews} (no think block, thinking off), Trace = \texttt{review\_with\_reasoning} (trained on teacher-rationale traces, thinking on). Criterion tags give the TTCW dimension (F fluency, Fl flexibility, O originality, E elaboration). QWK is quadratic weighted kappa against the synthetic teacher score on the rows that arm completed; bold marks the best of the six arms per criterion. R@1 is retrieval of the arm's own synthesized reference review within the criterion; each arm is
  scored on its own pool (no-trace pools 4{,}998 / 4{,}997 / 4{,}999; trace pools 4{,}355 / 4{,}542 / 4{,}295), and a smaller pool is an easier retrieval task, so R@1 is comparable down a column, not across arms. No metric here measures agreement with human judgment; no human annotation exists in this pipeline.}
  \label{tab:per_criterion_qwk_r1}
\end{table*}


\begin{table}[t]
\centering
\scriptsize
\setlength{\tabcolsep}{3pt}
\renewcommand{\arraystretch}{1.05}
\begin{tabular}{lccc}
\toprule
\textbf{Criterion} &
\textbf{Qwen3-8B} &
\textbf{Qwen3-4B} &
\textbf{Nemotron-8B} \\
\midrule
F1 Narrative Pacing      & 0.1132 & 0.1184 & 0.0496 \\
F2 Scene vs Exposition   & 0      & 0.0002 & 0.0002 \\
F3 Language Proficiency  & 0.0086 & 0.0129 & 0.0192 \\
F5 Understandability     & 0      & 0.0002 & 0 \\
Fl1 Perspective \& Voice  & 0.0005 & 0.0002 & 0.0002 \\
Fl2 Emotional Flexibility & 0.0002 & 0.0002 & 0.0006 \\
Fl3 Structural Flexibility& 0.0007 & 0      & 0 \\
O1 Originality: Theme    & 0.0005 & 0      & 0.0009 \\
O2 Originality: Thought  & 0.0002 & 0.0012 & 0.0011 \\
O3 Originality: Form     & 0      & 0      & 0.0004 \\
E1 World-Building        & 0.0034 & 0.0071 & 0.0069 \\
E2 Character Development & 0.0009 & 0.0005 & 0.0011 \\
\bottomrule
\end{tabular}
\caption{Loop hazard by TTCW criterion. F4 Narrative Ending and E3 Rhetorical Complexity are omitted because no loops are observed for either criterion in any of the three models.}
\label{tab:criterion_hazard}
\end{table}


\begin{table}[t]
\centering
\scriptsize
\setlength{\tabcolsep}{4pt}
\begin{tabular}{lccc}
\toprule
\textbf{Model pair} &
\textbf{Shared} &
\textbf{Same criterion} &
\textbf{Marginal exp.} \\
\midrule
Qwen3-8B / Qwen3-4B
& 315 & 97\% & 77\% \\
Qwen3-8B / Nemotron-8B
& 152 & 90\% & 58\% \\
Qwen3-4B / Nemotron-8B
& 155 & 86\% & 56\% \\
\bottomrule
\end{tabular}
\caption{Cross-family localisation of loop failures. \textit{Shared} denotes stories on which both models loop; \textit{Same criterion} is the proportion for which both loops occur in the same TTCW criterion. \textit{Marginal exp.} gives the agreement expected from the two models' marginal criterion distributions.}
\label{tab:cross_family_localisation}
\end{table}


\begin{table}[t]
\centering
\scriptsize
\setlength{\tabcolsep}{3.5pt}
\renewcommand{\arraystretch}{1.08}
\begin{tabular}{lccc}
\toprule
&
\textbf{Train} &
\textbf{Completed} &
\textbf{Failed} \\
\midrule
Trace words (p50)
& 11,315 & 11,584 & 22,599 \\

Max 20-gram repeat (p50)
& 1 & 1 & 815 \\

Max 20-gram repeat (p99)
& 14 & 3 & 4,139 \\

Hedge / 1k words
& 0.04 & 0.03 & 8.50 \\

\texttt{but} / 1k words
& 7.77 & 7.57 & 26.73 \\
\bottomrule
\end{tabular}
\caption{Distributional comparison between the 237{,}271 trace-supervised training targets and completed and failed generations. Completed generations remain close to the training distribution, whereas failures show substantially greater trace length, repetition, hedge frequency, and connective density.}
\label{tab:supervision_audit}
\end{table}


\begin{table}[t]
\centering
\scriptsize
\setlength{\tabcolsep}{4pt}
\renewcommand{\arraystretch}{1.08}
\begin{tabular}{lrc}
\toprule
\textbf{Evaluation subset} &
\textbf{$N$} &
\textbf{QWK} \\
\midrule
Trace-on, own completed
& 4,356 & 0.4945 \\

Trace-off, own completed
& 4,994 & 0.4489 \\

Trace-off on trace-on completed
& 4,350 & 0.4524 \\

Trace-off on trace-on failed
& 644 & 0.4219 \\

\midrule
Paired trace-on advantage
& 4,350 &
$+0.0420\,[+0.0341,+0.0495]$ \\
\bottomrule
\end{tabular}
\caption{Qwen3-8B decomposition of the effect of generating the supervised reasoning trace. The paired comparison uses the 4,350 stories completed under both inference modes and reports the QWK difference with its $95\%$ confidence interval.}
\label{tab:trace_value}
\end{table}

\clearpage
\section{Prompts}\label{sec:prompt}
Our dataset-construction prompts are adapted from the expert evaluation criteria of the Torrance Test of Creative Writing (TTCW) proposed by \citet{10.1145/3613904.3642731}. Since the metric definitions largely follow the original TTCW descriptions, we report only the final instruction pattern and scoring question used for each metric.

{\small
\setlength{\LTleft}{0pt}
\setlength{\LTright}{0pt}
\begin{longtable}{p{0.22\textwidth}p{0.74\textwidth}}
\caption{Final instruction patterns and scoring questions used in dataset construction. Full metric definitions follow the original TTCW criteria.}
\label{tab:prompt_templates}\\
\toprule
\textbf{Metric} & \textbf{Final prompt instruction} \\
\midrule
\endfirsthead

\toprule
\textbf{Metric} & \textbf{Final prompt instruction} \\
\midrule
\endhead

\midrule
\multicolumn{2}{r}{\textit{Continued on next page}} \\
\endfoot

\bottomrule
\endlastfoot

Narrative Pacing &
Given the story above, list out the scenes in the story in which time compression or time stretching is used, and argue for each whether it is successfully implemented. Then overall, give your reasoning about the question below and give an answer from 1 to 10, where 10 is the best score and 1 is the worst score.\\
& \textbf{Q)} How appropriate and balanced does the manipulation of time in terms of compression or stretching feel? \\
\midrule

Scene vs.\ Exposition &
Given the story above, answer the following question. Please first explain your reasoning step by step and then give an answer from 1 to 10, where 10 is the best score and 1 is the worst score.\\
& \textbf{Q)} How well does the story balance scene and summary/exposition, rather than relying heavily on one element? \\
\midrule

Language Proficiency \& Literary Devices &
Given the story above, please list out all the metaphors, idioms and literary allusions, and for each decide whether it is successful or whether it feels forced or too easy. Then overall, give your reasoning about the question below and give an answer from 1 to 10, where 10 is the best score and 1 is the worst score.\\
& \textbf{Q)} How sophisticatedly does the story use idiom, metaphor, or literary allusion? \\
\midrule

Narrative Ending Quality &
Given the story above, answer the following question. Please first explain your reasoning step by step and then give an answer from 1 to 10, where 10 is the best score and 1 is the worst score.\\
& \textbf{Q)} How natural and earned does the end of the story feel, rather than arbitrary or abrupt? \\
\midrule

Understandability \& Coherence &
Given the story above, answer the following question. Please first explain your reasoning step by step and then give an answer from 1 to 10, where 10 is the best score and 1 is the worst score.\\
& \textbf{Q)} How well do the different elements of the story work together to form a unified, engaging, and satisfying whole? \\
\midrule

Perspective \& Voice Flexibility &
Given the story above, answer the following question. Please first explain your reasoning step by step and then give an answer from 1 to 10, where 10 is the best score and 1 is the worst score.\\
& \textbf{Q)} How well does the story represent perspective and voice in a flexible and convincing way? \\
\midrule

Emotional Flexibility &
Given the story above, answer the following question. Please first explain your reasoning step by step and then give an answer from 1 to 10, where 10 is the best score and 1 is the worst score.\\
& \textbf{Q)} How well does the story achieve a balance between interiority and exteriority in a way that feels emotionally flexible? \\
\midrule

Structural Flexibility &
Given the story above, list each element in the story that is intended to be surprising. For each, decide whether the surprising element remains appropriate with respect to the entire story. Then overall, give your reasoning about the question below and give an answer from 1 to 10, where 10 is the best score and 1 is the worst score.\\
& \textbf{Q)} How well does the story contain turns that are both surprising and appropriate? \\
\midrule

Originality in Theme and Takeaway &
Given the story above, list out elements that are unique takeaways of this story for the reader. Then overall, give your reasoning about the question below and give an answer from 1 to 10, where 10 is the best score and 1 is the worst score.\\
& \textbf{Q)} How likely is it that an average reader of this story will obtain a unique and original idea from reading it? \\
\midrule

Originality in Thought &
Given the story above, are there any clichés in the story? If so, list out all the elements in this story that are cliché. Then overall, give your reasoning about whether the piece is negatively impacted by these clichés and give an answer from 1 to 10, where 10 is the best score and 1 is the worst score.\\
& \textbf{Q)} How original is the story as a piece of writing, without clichés? \\
\midrule

Originality in Form &
Given the story and the devices mentioned above, list each device used with a short explanation of whether it is successful or not. Then overall, give your reasoning about the question below and give an answer from 1 to 10, where 10 is the best score and 1 is the worst score.\\
& \textbf{Q)} How original is the story in its form? \\
\midrule

World-Building and Sensory Believability &
Given the story above, list out the elements in the story that call to each of the five senses. Then overall, give your reasoning about the question below and give an answer from 1 to 10, where 10 is the best score and 1 is the worst score.\\
& \textbf{Q)} How well does the writer make the fictional world believable at the sensory level? \\
\midrule

Character Development Depth &
Given the story above, list each character and the level of development. Then overall, give your reasoning about the question below and give an answer from 1 to 10, where 10 is the best score and 1 is the worst score.\\
& \textbf{Q)} How well does each character in the story feel developed at the appropriate complexity level, ensuring that no character is present merely to satisfy a plot requirement? \\
\midrule

Rhetorical Complexity &
Given the story above, answer the following question. Please first explain your reasoning step by step and then give an answer from 1 to 10, where 10 is the best score and 1 is the worst score.\\
& \textbf{Q)} How well do passages in the story involve subtext, and when subtext is present, how effectively does it enrich the story's setting rather than feel forced? \\

\end{longtable}
}
\end{document}